\documentclass[10pt,twocolumn,letterpaper]{article}

\usepackage[pagenumbers]{cvpr} 

\usepackage{graphicx}
\usepackage{amsmath}
\usepackage{amssymb}
\usepackage{booktabs}
\usepackage{xspace}
\usepackage{interval}
\usepackage{bm}
\usepackage{mathtools}
\usepackage{footmisc}

\usepackage{siunitx}
\usepackage{bbold}
\usepackage{multirow}
\usepackage{bbm}
\usepackage{amssymb}
\usepackage{bbding}
\usepackage[dvipsnames]{xcolor}
\usepackage{makecell}
\usepackage[inline]{enumitem}
\usepackage{algpseudocode,algorithm,algorithmicx}

\usepackage[pagebackref=true,breaklinks=true,letterpaper=true,colorlinks,citecolor=ForestGreen, bookmarks=false]{hyperref}

\usepackage[capitalize]{cleveref}
\crefname{section}{Sec.}{Secs.}
\Crefname{section}{Section}{Sections}
\Crefname{table}{Table}{Tables}
\crefname{table}{Tab.}{Tabs.}

\def\confName{CVPR}
\def\confYear{2022}

\begin{document}
	
	\title{2nd Place Solution to Google Universal Image Embedding}

	\author{Xiaolong Huang$^{1}$ \quad Qiankun Li$^{2,3}$\thanks{Corresponding Author: Qiankun Li (qklee@mail.ustc.edu.cn).}\\[2mm]
		$^{1}$School of Artificial Intelligence, Chongqing University of Technology,\\
		$^{2}$Institute of Intelligent Machines, Chinese Academy of Sciences,\\
		$^{3}$Department of Automation, University of Science and Technology of China
	}
	\maketitle
	
	\renewcommand{\thefootnote}{\fnsymbol{footnote}}
	
	\begin{abstract}
		
		This paper presents the 2nd place solution to the Google Universal Image Embedding Competition 2022. 
		We use the instance-level fine-grained image classification method to complete this competition. 
		We focus on data processing, model structure, and training strategies. 
		To balance the weights between classes, we employ sampling and resampling strategies. 
		For model selection, we choose the CLIP-ViT-H-14 model pretrained on LAION-2B.  Besides, we removed the projection layer to reduce the risk of overfitting. 
		In addition, dynamic margin and stratified learning rate training strategies also improve the model's performance. 
		Finally, the method scored 0.713 on the public leaderboard and 0.709 on the private leaderboard.  
		Code is available at \url{https://github.com/XL-H/GUIE-2nd-Place-Solution}.
		
	\end{abstract}
	
	\section{Introduction}
	\label{sec:intro}
	
	The Google Universal Image Embedding Competition \cite{2022google1} is part of the ECCV 2022 Instance-Level Recognition Workshop. Unlike previous competitions \cite{2021google1, 2020google2}, this year does not focus on the landmark domain. Instead, it requires that submitted models can be applied to many different object types. Specifically, participants were asked to submit a model that extracts a feature embedding of fewer than 64 dimensions for each image in the test set to represent its information. The submitted model should be able to retrieve the database images associated with a given query image. There are 200,000 index images and 5,000 query images in the test set, covering a variety of object types such as clothing, artwork, landmarks, furniture, packaging items, etc.
	
	In this paper, we present our 2nd place detailed solution to the Google Universal Image Embedding Competition 2022 \cite{2022google1}. 
	Since the competition did not provide training data, building a generic dataset was important. 
	According to the official categories, we sampled from various open benchmarks and preprocessed the data to build a universal dataset containing 7,400,000 images. 
	For image embedding, standard solutions include Contrastive Learning, Generic Classification, Fine-grained Classification, and so on.  
	Considering that the test set for the competition is annotated at the instance level, we use a Fine-grained Classification.
	To obtain a strong baseline, the Transformer-based\cite{swintrans, swintransv2, hyb-swin} OpenClip-ViT-H-14 \cite{2021Robust} model is used for feature extraction. 
	To enhance the stability of model training, we use pre-trained weights on a two-billion-scale image-text pairs dataset LAION-2B \cite{2021LAION} to initialize the OpenCLIP \cite{2021Robust}. 
	In addition, dynamic margin \cite{arcface} and stratified learning rate training strategies also helped us win 2nd place in the competition.
	
	\begin{figure*}
		\centering
		\includegraphics[width=0.8\linewidth]{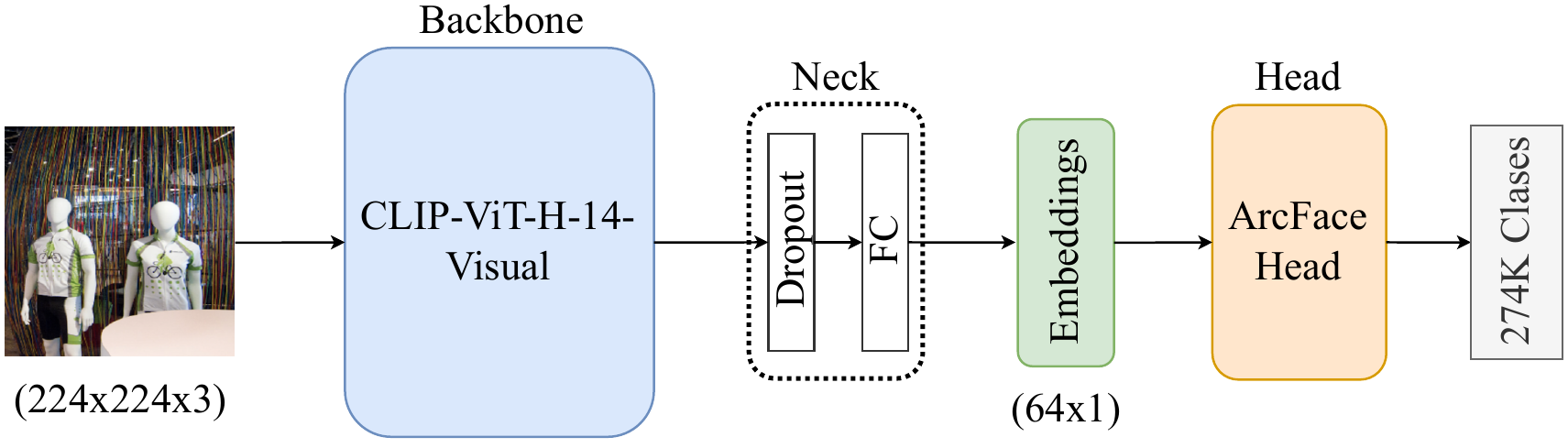}
		\caption{\label{fig:1} Overview of the moodel.}	
	\end{figure*}

	\section{Method}
	\subsection{Datasets}
	\vspace{0.05in}
	\noindent\textbf{Datasets selection.}
	Since we use the Fine-grained Classification method, the corresponding datasets should select instance-level data. 
	After the information collected from major open websites, the selected datasets are as follows \cite{datasets}: Aliproducts, Art MET, DeepFashion (consumer-to-shop), DeepFashion2(hard-triplets), Fashion200K, ICCV 2021 LargeFineFoodAI, Food Recognition 2022, JD Products 10K, Landmark2021, Grocery Store, rp2k, Shopee, Stanford Cars, Stanford Products. 
	The tasks of these datasets are related to image retrieval or fine-grained recognition.
	
	\vspace{0.05in}
	\noindent\textbf{Datasets pre-processing.}
	The ability of the model to identify a class of instances is not directly related to the number of images of the class. 
	Therefore, the reasonable size of each class may better improve the model performance and save training time. 
	We sampled 100 images from categories with more than 100 images and filtered out categories with less than 3 images. To ensure the model can extract enough information from afferent instances in the training data, we resampled lightly to balance the class weight. The minimum number of each class was set to 20. Finally, we obtained 7,400,000 images and then stratified 5\% of them with K-fold for validation.
	
	\subsection{Model Structure}
	Fig.~\ref{fig:1} is the overview of our model. 
	The model used OpenCLIP-ViT-H-14-Visual \cite{2021Robust} as the backbone, a fully connected layer with the 0.2 rate dropout as the neck, and ArcFace \cite{arcface} as the head. 
	Since the parameter values of project layers in OpenCLIP-ViT-H-14 \cite{2021Robust} are large-scale, the direct porting of CLIP as the feature extraction backbone will easily cause overfitting.  
	To solve this problem, we can normalize the feature embedding of the backbone output or remove the projection layer of the backbone. Since the latter can also reduce the computational costs of the model, we choose the latter. 
	
	\subsection{Training Skills}
	\vspace{0.05in}
	\noindent\textbf{Dynamic margin.}
	We found that the slightly changing margin of ArcFace \cite{arcface} head influenced the score on the public leaderboard significantly. Therefore, We implement a dynamic margin \cite{arcface} strategy to avoid divergence during the first few training epochs. The margin value is defined as:
	\begin{equation}
	{m_e}= {min [m_{init} + s*(e-1), m_{max})]},
	\label{eq:me}
	\end{equation}
	where $m_e$, $m_{init}$, and $m_{max}$ represent the current epoch of margin, initial margin, and maximum margin. The $s$ represent the stride.
	
	\vspace{0.05in}
	\noindent\textbf{Stratified learning rate.}
	After training in contrast learning, OpenCLIP-ViT-H-14-Visual \cite{2021Robust} has built a massive space of hidden information for instance-level perception. 
	However, Since the competition is highly correlated with instance-level image retrieval, it is still slightly different from the expected results of the model. 
	To further improve the score, we need a way to transfer the hidden information from the contrast learning model to the instance-level retrieval model as harmlessly as possible. 
	Therefore, we implemented an effective hierarchical learning strategy when fine-tuning the model. 
	Specifically, we set a normal-sized learning rate for the non-backbone and a lower learning rate for the backbone. 
	The magnitude of the two learning rates are defined as: 
	\begin{equation}
	{lr_b} = {lr * c},
	\label{eq:lr}
	\end{equation}
	where  $ lr_b $  and  $lr$  represent the learning rate for the backbone and learning rate of the non-backbone. The $c$ represent the reduction factor, and  $0<c<1 $.
	
	\section{Experiments}
	In this section, the implementation Details and detailed experimental results are introduced. 

	\subsection{Implementation Details}
	All experiments are based on Pytorch 1.9 framework, and 6 * NVIDIA A40 were used for training. 
	We set the input image size as $224 \times 224$, 
	The following data augmentation methods from the Albumentaions \cite{buslaev2020albumentations} library were used: HorizontalFlip, RandomBrightnessContrast, ShiftScaleRotate, and CoarseDrouout. 
	We employed the AdamW optimizer and Cosine Annealing learning rate schedule with an initial learning rate of 1e-4. 
	All models have trained no more than 15 epochs. 
	Model ViT-H-14 used in our final solution was only trained for three epochs, two epochs fine-tuned with unfrozen backbone, then frozen backbone to fine-tune for another one epoch.
	
	\begin{table}\small
		\centering
		\caption{Comparison of different backbones. $^*$ represents from OpenAI \cite{2021LAION}, $^\#$ represents from OpenCLIP \cite{2021Robust}.}
		\renewcommand{\tabcolsep}{5pt}
		\begin{tabular}{lccc}
			\toprule
			Backbone & Image Size & Public Score & Private Score \\ 
			\midrule
			ViT-L-14$^*$ & $224\times224$ & 0.579 & 0.587 \\ 
			ViT-L-14-336$^*$ & $336\times336$ & 0.598 & 0.611 \\ 
			ViT-L-14$^\#$ & $224\times224$ & 0.636 & 0.646 \\ 
			ViT-g-14$^\#$ & $224\times224$ & 0.647 & 0.656 \\ 
			ViT-H-14$^\#$ & $224\times224$ & 0.662 & 0.666 \\ 
			\bottomrule
		\end{tabular}
		\label{tab:1}
	\end{table}

	\begin{table*}[t]\small
		\centering
		\caption{Results of Ablation study.}
		\renewcommand\tabcolsep{5pt}
		\begin{tabular}{lcccccccc} 
			\toprule 
			Backbone  & Data Scale& Freeze Backbone & Dynamic Margin& Remove Proi Layer& Stratified LR& Flip & Public & Private \\
			\midrule
			ViT-H-14 &  1.1 M & \checkmark& &  & & & 0.656 & 0.661 \\ 
			ViT-H-14 &  1.1 M &\checkmark&  &\checkmark  & & & 0.662 & 0.666 \\ 
			ViT-H-14 & 1.1 M & \checkmark & \checkmark  &\checkmark  & & & 0.670 & 0.670 \\ 
			ViT-H-14 & 1.1 M &&\checkmark  &\checkmark  &\checkmark & & 0.697 & 0.701 \\ 
			ViT-H-14 & 2.5 M &&\checkmark  &\checkmark  &\checkmark & & 0.702 & 0.701 \\ 
			ViT-H-14 & 7.4 M &&\checkmark  &\checkmark  &\checkmark & & 0.707 & 0.709 \\ 
			ViT-H-14 & 7.4 M &&\checkmark  &\checkmark  &\checkmark &\checkmark & 0.713 & 0.709 \\ 
		\end{tabular}
		\label{tab:2}
	\end{table*}

	\subsection{Evaluation Metric}
	Submissions are evaluated according to the mean $Precision @ 5$ metric, introducing a small modification to avoid penalizing queries with fewer than 5 expected index images. 
	In detail, the metric is computed as follows:
	\begin{equation}
	m P @ 5=\frac{1}{Q} \sum_{q=1}^{Q} \frac{1}{\min \left(n_{q}, 5\right)} \sum_{j=1}^{\min \left(n_{q}, 5\right)} \operatorname{rel}_{q}(j),
	\label{eq:metric}
	\end{equation}
	where, $Q$ is the number of query images. $n_q$ is the number of index images containing an object in common with the query image, and that $n_q > 0$. ${rel}_{q}(j)$ denotes the relevance of prediciton $j$ for the $q$-th query: it’s 1 if the $j$-th prediction is correct, and 0 otherwise

	\subsection{Comparison of Backbones}
	We tried various backbones in the competition, all of which came from OpenAI \cite{2021LAION} and OpenCLIP \cite{2021Robust}. 
	When comparing the performance of each backbone network, we uniformly remove the projection layer and freeze weight parameters.
	As listed in Table \ref{tab:1}, all models achieved good results on the leaderboard, among which ViT-H-14$^\#$ has the highest score on the leaderboard.
	
	\subsection{Ablation Study}
	We compared the different strategies based mainly on ViT-H-14 scores on the public leaderboard. And Fully connected layer with dropout rate 0.2 as neck, ArcFace \cite{arcface} as the head. $s$ of Dynamic Margin in Eq. (\ref{eq:me}) is set to $1$, $c$ of Stratified LR in Eq. (\ref{eq:lr}) is set to 1e-3.
	Table \ref{tab:2} listed the most representative strategies.
	
	\section{Conclusion}
	In this paper, we present our 2nd place solution detailedly. We used sampling and resampling to balance class weights, redesigned the backbone structure to reduce the risk of overfitting, and implemented dynamic margin and stratified learning rate training strategies to boost our score significantly.
	
{\small
	\bibliographystyle{ieee_fullname}
	\bibliography{egbib}
}

\end{document}